\documentclass[10pt,twocolumn,letterpaper]{article}

\usepackage{cvpr}
\usepackage{times}
\usepackage{epsfig}
\usepackage{graphicx}
\usepackage{amsmath}
\usepackage{amssymb}
\usepackage{booktabs,subfig} %
\usepackage{multirow}
\usepackage{bm}
\usepackage{tabularx}
\usepackage{threeparttable}
\usepackage[dvipsnames]{xcolor}
\usepackage{cite}
\usepackage{url}

\usepackage[pagebackref=true,breaklinks=true,letterpaper=true,colorlinks,bookmarks=false]{hyperref}

\makeatletter
\def\hlinew#1{%
  \noalign{\ifnum0=`}\fi\hrule \@height #1 \futurelet
   \reserved@a\@xhline}
\makeatother

\newcommand{\specialcell}[2][c]{%
  \begin{tabular}[#1]{@{}c@{}}#2\end{tabular}}

\cvprfinalcopy %

\def\cvprPaperID{4018} %

\begin{document}

\title{Leveraging Crowdsourced GPS Data for Road Extraction from Aerial Imagery}

\author{Tao Sun, \quad Zonglin Di, \quad Pengyu Che, \quad Chun Liu, \quad Yin Wang\\
Tongji University, Shanghai, China\\
{\tt\small \{suntao, dizonglin, chepengyu, liuchun, yinw\}@tongji.edu.cn}
}

\maketitle
\begin{abstract}
Deep learning is revolutionizing the mapping industry.
Under lightweight human curation, computer has generated almost half of the roads in Thailand on OpenStreetMap~(OSM) using high resolution aerial imagery.
Bing maps are displaying 125~million computer generated building polygons in the U.S.
While tremendously more efficient than manual mapping, one cannot map out everything from the air.
Especially for roads, a small prediction gap by image occlusion renders the entire road useless for routing.
Misconnections can be more dangerous.
Therefore computer based mapping often requires local verifications, which is still labor intensive.
In this paper, we propose to leverage crowd sourced GPS data to improve and support road extraction from aerial imagery.
Through novel data augmentation, GPS rendering, and 1D transpose convolution techniques, we show almost 5\% improvements over previous competition winning models, and much better robustness when predicting new areas without any new training data or domain adaptation.
\end{abstract}

\section{Introduction}\label{sec:intro}
Segmentation of aerial imagery has been an active research area for more than two decades~\cite{mayer2006test,aksoy2008performance}.
It is also one of the earliest applications of deep convolutional neural nets~(CNN)~\cite{mnih2012learning}.
Today, using deep convolutional neural nets over high resolution satellite imagery, Facebook has added 370 thousand km of computer generated roads to OpenStreetMap~(OSM)~\cite{OSM} Thailand, accounting for 46~\% of the total roads in the country, which is on display for all Facebook users~\cite{OSM_AI_assisted,STOM:Facebook:2018}.
Microsoft used similar techniques to add 125 million building polygons to Bing maps U.S., five times more than those on OSM~\cite{STOM:Microsoft:2018}.

\begin{figure}[!h]
\centering
\subfloat[Occlusions by trees, buildings, and shadows are challenging without GPS\label{fig:gap:a}]{
\includegraphics[width=0.49\columnwidth]{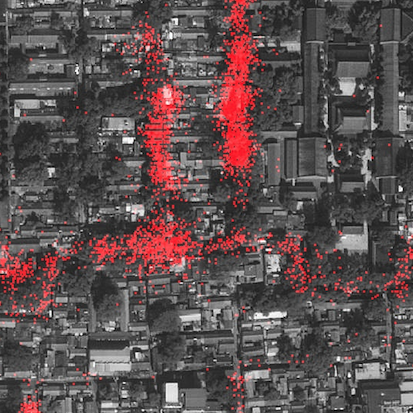}\hspace{0.01\columnwidth}
\includegraphics[width=0.49\columnwidth]{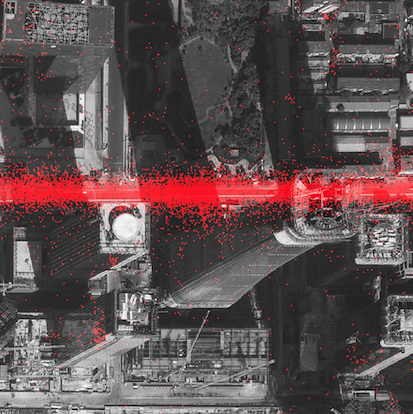}
}\\
\vspace{-1ex}
\subfloat[Roads susceptible to over connection in post-processing without GPS\label{fig:gap:b}]{
\includegraphics[width=0.49\columnwidth]{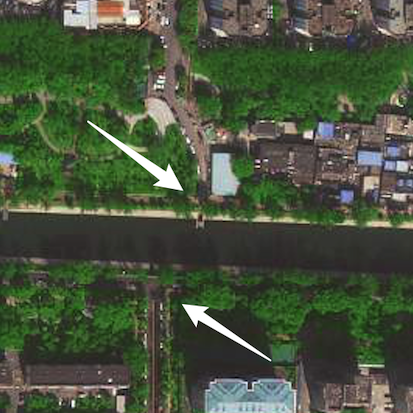}\hspace{0.01\columnwidth}
\includegraphics[width=0.49\columnwidth]{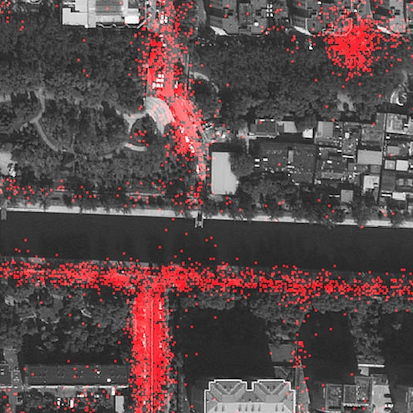}
}
\caption{Crowdsourced GPS data helps road extraction when aerial imagery alone is insufficient or challenging. Here each red dot represents a taxi GPS sample.}
\label{fig:gps-benefit}
\vspace{-1ex}
\end{figure}

Despite real-world applications, mapping by aerial imagery has its limitations.
The top challenge is overfitting.
The deep neural net models often deteriorate miserably with new terrain, new building styles, new image styles, or new resolutions.
Other than the model limitation, occlusions by vegetation, buildings, and shadows can be excessive.
Many features are indistinguishable from the air, e.g., dirt roads and bare fields, cement pavements and building tops, alleys in slum areas.
Bad weather, low satellite angle, and low light angle further complicate the issue.
Even if the feature is perfectly clear, mapping often needs local knowledge.
Trails and roads may have same appearances.
Houses and storage sheds may have similar sizes and roofs.
To make things worse, mapping has low tolerance for errors.
Especially for roads, incorrect routes cause longer travel time, lead people to restricted areas, and even cause fatal accidents~\cite{Wang:2013}.
Because of these reasons, OSM prefers local mappers for each area, and even requires local verification for large-scale edits~\cite{OSM_AI_assisted}.

With a smart phone or any other GPS device, one can easily travel a street and verify its existence with the recorded trace.
Going through all streets systematically and regularly for updates, however, is a labor intensive job that is costly and error prone.
On the other hand, crowdsourced GPS data are much cheaper and increasingly abundant~\cite{liu2012mining,gis/BiagioniE12,Wang:2013,shan2015cobweb,karagiorgou2017layered}.
Figure~\ref{fig:gps-benefit} illustrates how crowdsourced GPS data, albeit noisy, can help discover roads, confirm road continuity, and avoid misconnection.

In this paper, we propose to fuse crowdsourced GPS data with aerial imagery for road extraction.
Through large taxi and bus GPS datasets from Beijing and Shanghai, we show that crowdsourced GPS data has excessive noise in both variation and bias, and high degrees of disparity in density, resolution, and distribution. 
By rendering the GPS data as new input layers along with RGB channels in the segmentation network, together with our novel GPS data augmentation techniques and 1D transpose convolution, our model significantly outperforms existing models using images or GPS data alone.
Our data augmentation is especially effective against overfitting.
When predicting a new area, the performance drop is much less than the model with image input only, despite completely different GPS data quantity and resolution.
We will publish the code\footnote{\href{https://github.com/suniique/}{https://github.com/suniique/}} and our data is available upon request.

\section{Related Work}

Aerial imagery segmentation has been a very active research area for a long time.
We refer readers to some performance studies and references therein for early algorithms~\cite{mayer2006test,aksoy2008performance}.
Like many other image processing problems, these early solutions are often limited in accuracy and difficult to generalize to real-world datasets.

Mnih first used a deep convolutional neural net similar to LeNet~\cite{lecun1998gradient} to extract roads and buildings from a 1.2~m/pixel dataset in the U.S.~\cite{mnih2012learning,MnihThesis}.
Moving to developing countries with more diversified roads and buildings, Facebook showed that deeper neural nets perform much better on a 50~cm/pixel dataset~\cite{Wang:2016keynote}.
Both of these early approaches convert the semantic segmentation problem into a classification problem by classifying each pixel of a center square, e.g., 32 x 32, as road or building from a larger image patch, e.g., 128 x 128. 
Stitching these center squares together is the final output for a large input image.
Performance issues aside, this classification approach cannot learn complicated structures such as street blocks and building blocks due to limited input size.

With the commercial availability of 30~cm/pixel satellite imagery and low-cost aerial photography drones,
more public high-resolution datasets become available~\cite{wang2017torontocity,huang2018large,yokoya2018open,xia2018dota,DeepGlobe18}.
These new datasets and industrial interests lead to a proliferation of research activities recently~\cite{volpi2017dense,mattyus2017deeproadmapper,bastani2018roadtracer,zhu2017deep}.
Semantic segmentation models based on the fully convolutional neural net architecture become main stream~\cite{shelhamer2017fully}.
In a recent challenge~\cite{DeepGlobe18}, all top solutions used variants of U-net~\cite{ronneberger2015u} or Deeplab~\cite{chen2018deeplab} to segment an entire image at once, up to 1024 x 1024 pixels.
A larger input size gives more context, which often leads to more structured and accurate prediction results.

With new models and multi-country scale datasets, many real-world applications emerge.
Most notably, Facebook has recently added 370 thousand km of roads extracted from satellite imagery to OSM~\cite{OSM} Thailand, or 46~\% of the total roads in the country~\cite{OSM_AI_assisted,STOM:Facebook:2018}.
Microsoft is displaying 125 million computer generated building polygons on Bing US maps, in contrast to the 23 million polygons from OSM also on display that are mostly manually created or imported~\cite{STOM:Microsoft:2018}.

Comparing to other computer vision applications, road mapping has little margin for error.
Prediction gaps make the entire road useless for routing, and therefore have attracted lots of attention.
Mnih noticed the problem early on and used Conditional Random Fields in post-processing to link broken roads~\cite{MnihThesis}.
Another popular technique to link roads is shortest path search~\cite{Wang:2016keynote,mattyus2017deeproadmapper}.
Line Integral Convolution can smooth out broken roads in post-processing too~\cite{li2016road}.
More recent works try to address the problem in prediction instead of post-processing, e.g., through a topology-aware loss function~\cite{mosinska2018beyond} or through an iterative search process guided by CNNs~\cite{bastani2018roadtracer}.
We must be careful to link roads because incorrect connections are more dangerous than missing connections in routing.
Our approach complements the above mentioned methods because GPS data can confirm the connectivity or the absence of it regardless of image occlusion or other issues.

Road inferencing from GPS traces has been studied for a long time too~\cite{kdd/RogersLW99,SchrodlWRLW04,pervasive/DaviesBH06}.
Most early works use dense GPS samples from controlled experiments.
Recent works explored crowdsourced GPS data under various sampling interval and noise levels~\cite{liu2012mining,gis/BiagioniE12,Wang:2013,shan2015cobweb,karagiorgou2017layered}.
Kernel Density Estimation is a popular method robust against GPS noise and disparity~\cite{pervasive/DaviesBH06,liu2012mining,gis/BiagioniE12}.

There is limited research work using both GPS data and aerial imagery.
One idea filters out GPS noise by road segmentation before road inferencing~\cite{yuan2016image}.
Our preliminary work explored the idea of rendering GPS data as a new CNN input layer, but the segmentation model used was a bit outdated and the GPS data was from a controlled experiment~\cite{sun2018combining}.
This paper experiments with many state-of-the-art segmentation models and crowdsourced GPS datasets several orders of magnitude bigger and noisier.
\section{Crowdsourced GPS Data} \label{sec:gps}
\begin{figure}[!t]
\centering
\subfloat[Excessive noise\label{fig:gps:a}]{
\includegraphics[width=0.49\columnwidth]{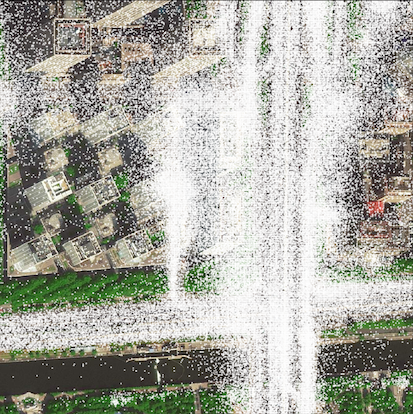}
}
\subfloat[Waiting areas\label{fig:gps:b}]{
\includegraphics[width=0.49\columnwidth]{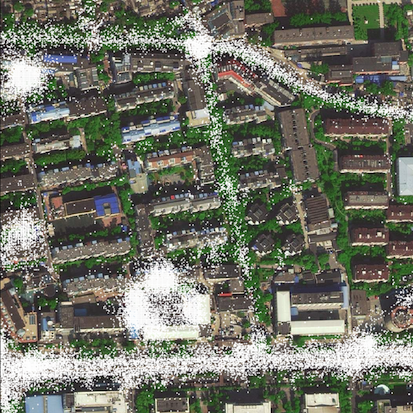}
}\\
\subfloat[Misalignment\label{fig:gps:c}]{
\includegraphics[width=0.49\columnwidth]{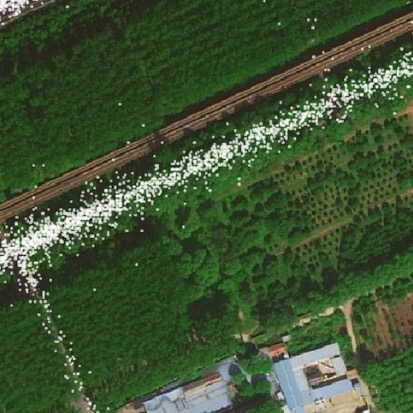}
}
\subfloat[Outdated data\label{fig:gps:d}]{
\includegraphics[width=0.49\columnwidth]{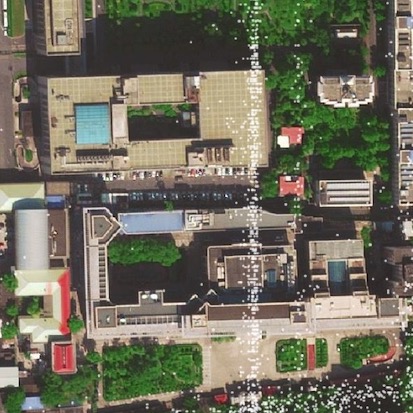}
}
\caption{Typical issues with crowdsourced GPS data}
\label{fig:gps-noise}
\vspace{-1ex}
\end{figure}

We collected two taxi and bus GPS datasets from Beijing and Shanghai, respectively.
The Beijing dataset is about one week of data with around 28 thousand taxis and 81 million samples.
The Shanghai dataset spans about half an year with around 16 thousand taxis and 1.5 billion samples.
In both cases, each sample includes a timestamp, latitude, longitude, speed, bearing, and taxi status flags. 
Although taxis have different behaviors and trajectories than other GPS data sources, we believe many characteristics and issues in our datasets are quite representative.
Therefore our method applies to other datasets.

Under ideal conditions, GPS samples follow a 2D Gaussian distribution~\cite{Diggelen07}.
Predicting roads can be straightforward if the samples are dense and evenly distributed.
In practice, multipath errors occur in urban canyons, inside tunnels, and under elevated highways or bridges. 
GPS receivers vary in quality and resolution, and may integrate Kalman filters that are not Gaussian.
Some datasets purposefully reduce resolution and/or add random noise for privacy protection.
Figure~\ref{fig:gps:a} is an example of noisy GPS samples mainly due to urban canyon and elevated roads.

Even if the samples are perfectly Gaussian distributed, unlike controlled experiments or surveys, crowdsourced GPS data are not evenly distributed along each road.
Highways and intersections can have orders of magnitude more data than other road areas.
Some residential roads are not traveled at all.
Depending on the source of data, there may be concentrations of samples in non-road areas.
Figure~\ref{fig:gps:b} shows three high density clusters outside of artery roads, likely popular taxi waiting areas.
Misalignment can occur with shifted data, e.g., Fig.~\ref{fig:gps:c}, or with different time periods when the data are taken, e.g., Fig.~\ref{fig:gps:d}.

\begin{figure}[!htbp]
    \subfloat[Beijing sampling interval (s)\label{fig:bj-sample-interval}]
    {\includegraphics[width=0.49\columnwidth]{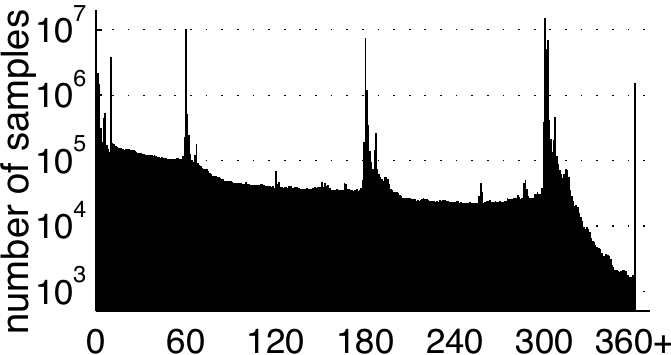}}
    \subfloat[Beijing sample speed (km/h)\label{fig:bj-sample-speed}]
    {\includegraphics[width=0.49\columnwidth]{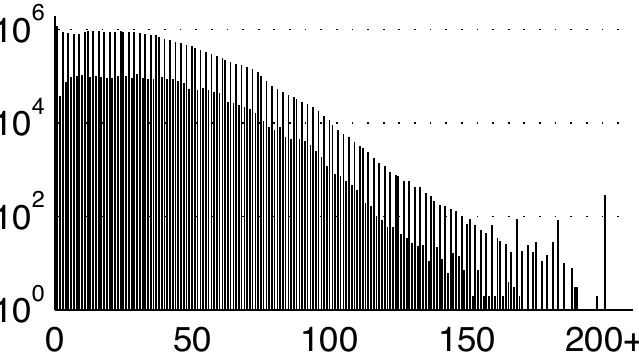}}\\
    \subfloat[Shanghai sampling interval (s)\label{fig:sh-sample-interval}]
    {\includegraphics[width=0.49\columnwidth]{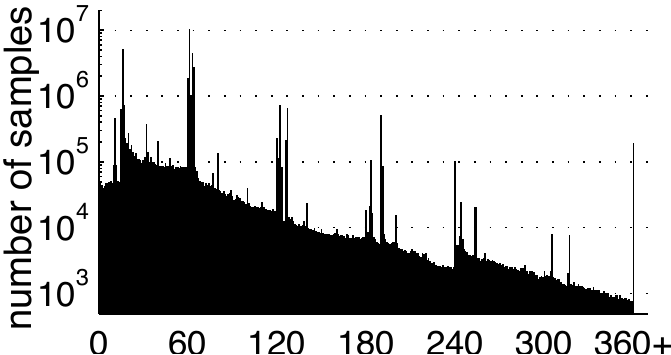}}
    \subfloat[Shanghai sample speed (km/h)\label{fig:sh-sample-speed}]
    {\includegraphics[width=0.49\columnwidth]{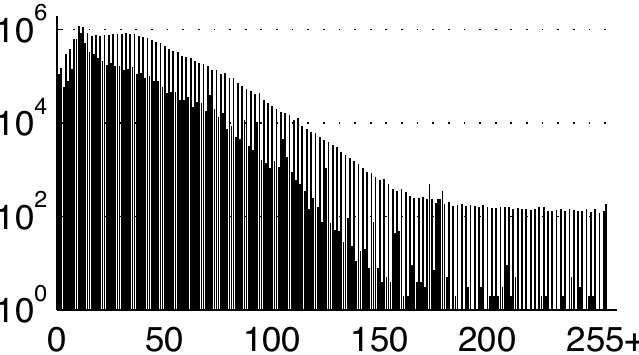}}
    \caption{Distributions of sampling interval and speed}%
    \label{fig:sample-hist}
\end{figure}

\begin{table}[!htbp]
\center
\small
  \caption{Typical measurement resolutions in our datasets}
  \label{tab:dataresolution}
  \begin{tabularx}{\columnwidth}{|X|cc|}
    \hline
    \multirow{2}{*}{Resolution} & \multicolumn{2}{c|}{Dataset} \\
    \cline{2-3}
     & Beijing & Shanghai \\
    \hline
	\hline
	lat/lon (degree) & 1/100,000 & 1/60,000 or 1/10,000\\
	speed (km/h) & 1 or 2 & 1 or 2 \\
	bearing (degree) & 3 or 10 & 2 or 45\\
    \hline
  \end{tabularx}
\end{table}

Different vehicles may use different GPS receivers with different settings.
Figure~\ref{fig:sample-hist} shows the log scale distributions of sampling intervals and device-measured speed of our datasets.
It is obvious from the figure that different taxis have different sampling interval settings, most notably at 10, 60, 180, and 300 seconds for the Beijing dataset, and 16 and 61 seconds for the Shanghai dataset.
The speed distribution shows two layers of outline curves because the samples have different speed resolutions, most commonly 1 and 2~km/h.
Therefore the outer layer corresponds to even numbers and the inner layer corresponds to odd numbers.
Latitude, longitude, and bearing have different resolutions too, summarized in Table~\ref{tab:dataresolution}.
Most Beijing taxis are at $10^{-5}$ degree, or roughly 1~m.
Shanghai taxis have resolutions as low as $10^{-4}$ degree, or roughly 10~m.
Our satellite imagery has a resolution of 50~cm/pixel that is higher than our GPS data.
Therefore, there is the mosaic effect where some pixels have no GPS samples and some pixels may have multiple samples as the data quantity increases; see Fig.~\ref{fig:gps-noise} zoomed in.
Crowdsourced GPS data are cheap and abundant.
There can be multiple datasets for just one area.
We must develop a model robust against different data characteristics so there is no need to retrain the model with new datasets.

\section{Method}

By rendering GPS data as new input layers like RGB channels, our method applies to all existing CNN-based semantic segmentation networks.
GPS data augmentation prevents overfitting and gives a robust model against different GPS data characteristics.
Replacing the 3$\times$3 transpose convolution in the decoder by 1D transpose convolution gives better accuracy, called 1D decoder for the rest of this paper.

\subsection{Architecture}
\begin{figure}[!htbp]
\center
    \subfloat[Overview\label{fig:arch}]
	{\includegraphics[width=1\columnwidth]{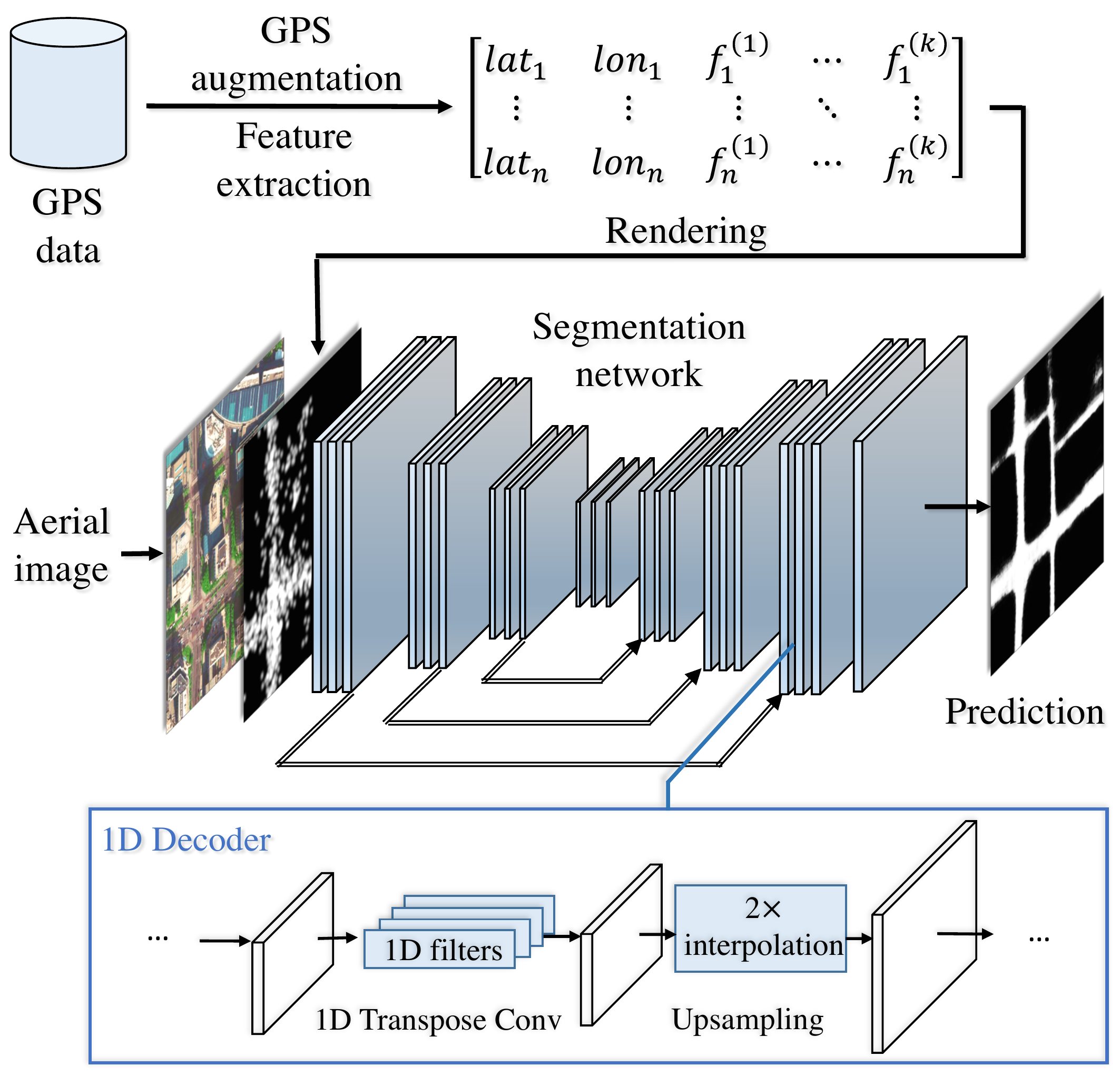}}\\
	\subfloat[1D transpose convolution block\label{fig:1dconv}]
	{\includegraphics[height=3.6cm]{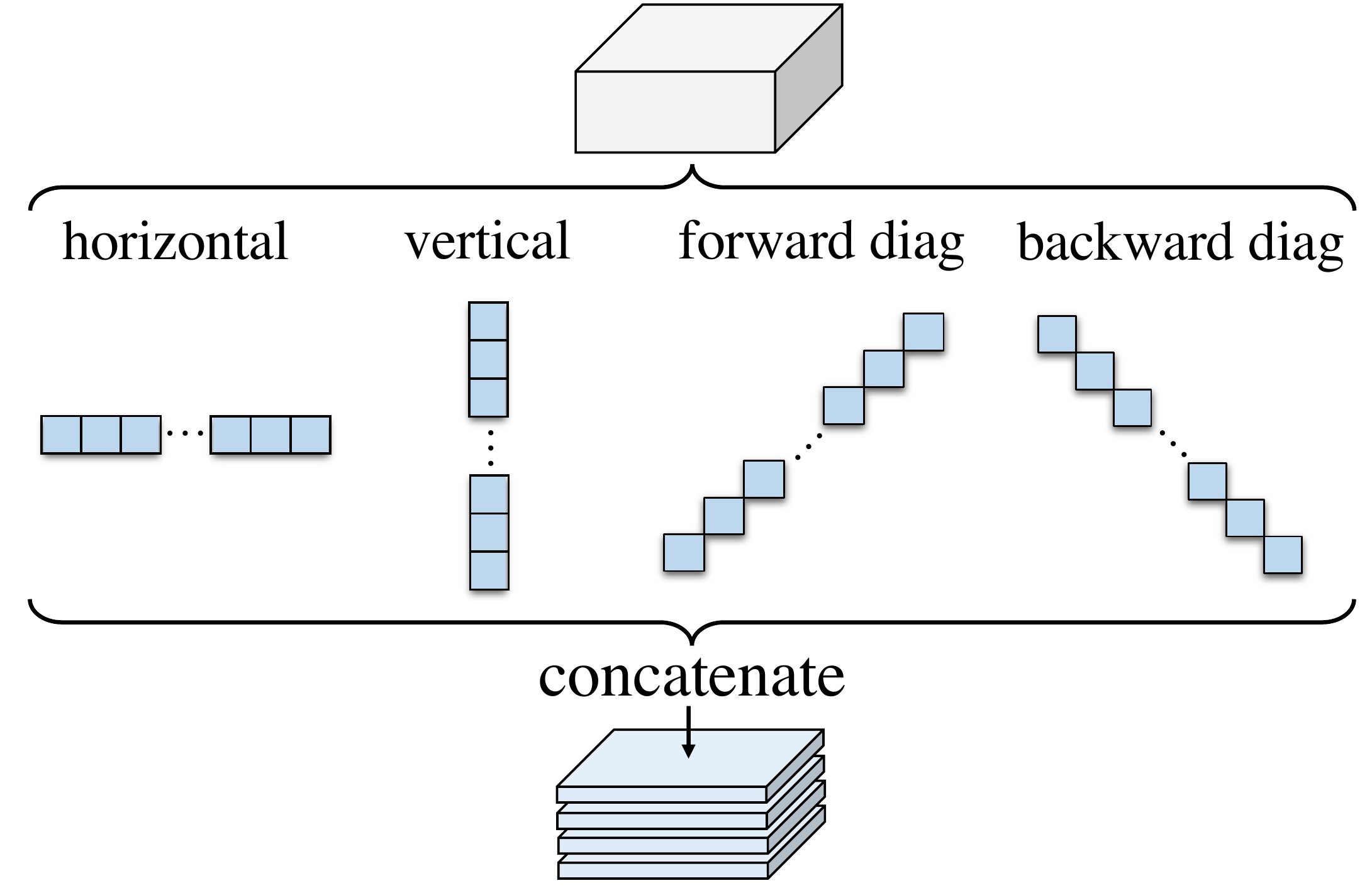}}
\caption{Network architecture}
\end{figure}

In DeepGlobe'18 road extraction challenge~\cite{DeepGlobe18}, all top teams used variants of fully convolutional net for pixel segmentation~\cite{shelhamer2017fully}, e.g., U-Net~\cite{ronneberger2015u} and DeepLab~\cite{chen2018deeplab}.
The winner team modified LinkNet~\cite{chaurasia2017linknet} that is very similar to U-net, by adding dilated convolutions to accommodate much larger input size and to produce more structured output, called D-LinkNet~\cite{zhou2018d}.
We propose to render GPS data as images and concatenate with RGB channels as the input to the segmentation net; see Fig.~\ref{fig:arch}.
Therefore our method applies to most existing segmentation networks. 
More specifically, based on the input image coordinates, we query database for the relevant GPS data in the area and get for example $n$ samples where each sample $i$ has coordinates $lat_i, lon_i$ and other features like sampling interval and vehicle speed $f_i^{(1)}, ... f_i^{(k)}$.
Like image augmentation frequently employed in image processing training, we augment GPS data to prevent overfitting.
Afterwards, we render the data as one or multiple image layers based on the number of features used.

Unlike natural objects, roads are thin, long, and often straight.
The square kernels that dominate most CNN architectures have square receptive fields that are more suitable for natural objects of bulk shapes.
For roads, it takes a very large square to cover a long straight road, where many pixels can be irrelevant.
The 1D filters are more aligned with road shapes.
We find that these 1D filters are most effective in the decoder block as replacements for 3$\times$3 transpose convolutions, as the lower portion of Fig.~\ref{fig:arch} depicts.

Let $\bm{k} \in \mathbb{R}^{2r+1}$ denotes the 1D transpose convolution filter of size $2r+1$, and $\bm{y_I} \in \mathbb{R}^{H\times W}$ be the result of 1D transpose convolution of input $\bm{x}\in \mathbb{R}^{H\times W}$ and the filter $\bm{k}$ at direction $\bm{I}=(I_h, I_w)$.
We have
\begin{equation}
\begin{aligned}
\bm{y_I}[i, j] & = (\bm{x} *^T \bm{k})_{\bm{I}} = \\
& \sum_{t=-r}^{r} \bm{x}[i + I_h t , j + I_w t] \cdot \bm{k}[r - t]
\end{aligned}
\end{equation}
where $\bm{x} *^T \bm{k}$ is the transpose convolution operation, and $\bm{I}$ is the direction indicator vector of the 1D filter, which takes four values $(0,1),(1,0),(1,1),(-1,1)$ for horizontal, vertical, forward diagonal, and backward diagonal transpose convolution, respectively, shown in Fig.~\ref{fig:1dconv}.

We set $r=4$ and thus each 1D filter has 9 parameters, the same as the 3$\times$3 transpose convolution filter.
Our 1D decoder replaces each of the 3$\times$3 transpose convolution layer by four sets of 1D filters of the four directions in concatenation.
The number of 1D filters in each set is 1/4 of the total number of 3$\times$3 filters.
Therefore, the total number of network parameters and the computation cost remain the same.
Our 1D decoder is especially effective against roads with sparse GPS samples, e.g., residential roads, by reducing gaps in the prediction.

\subsection{Data Augmentation}
\begin{figure}[!hbtp]
\center
 \vspace{-4ex} 
\subfloat[Original data\label{fig:aug:original}]{
\includegraphics[width=0.49\columnwidth]{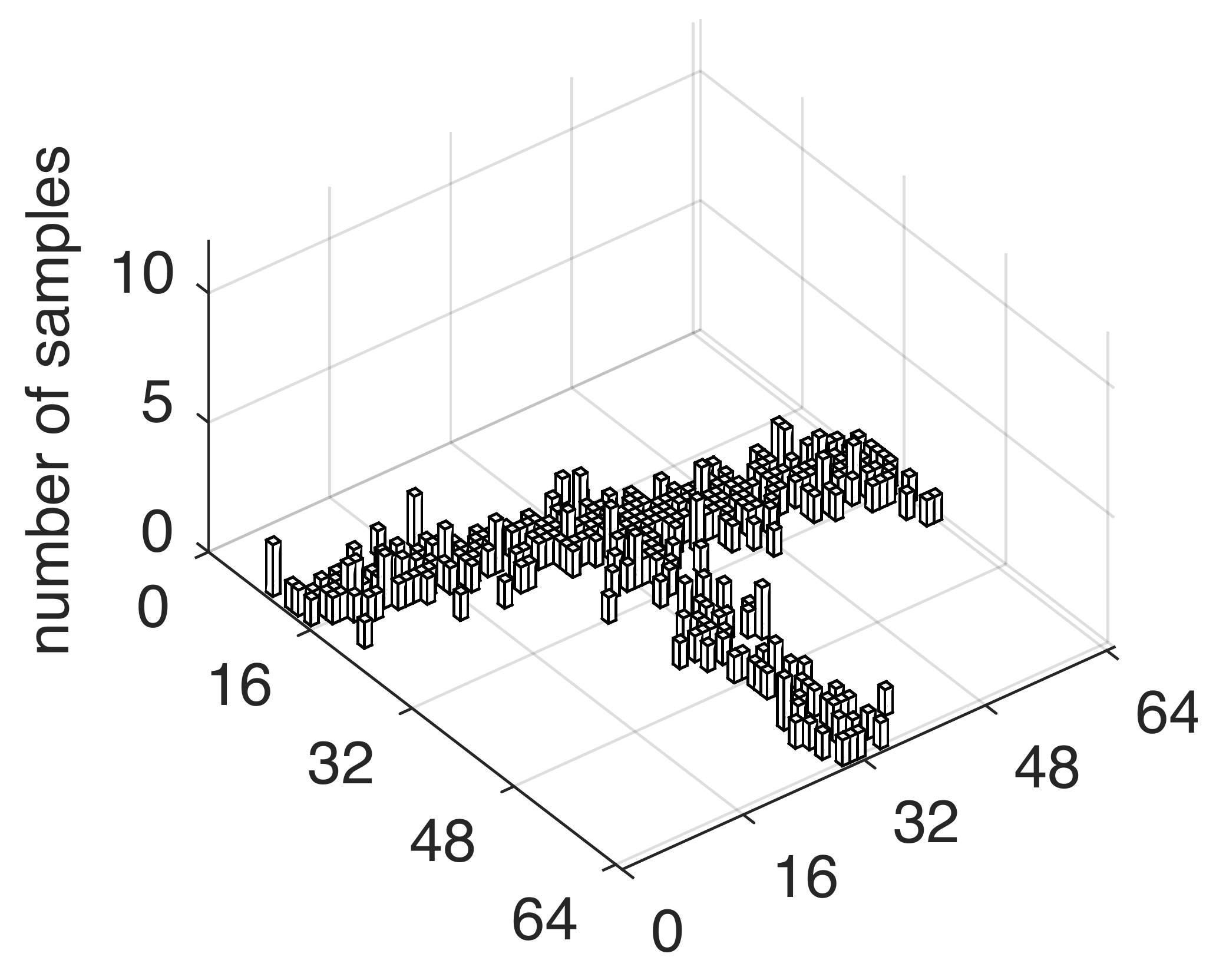}
}
\subfloat[Subsampling\label{fig:aug:subsample}]{
\includegraphics[width=0.49\columnwidth]{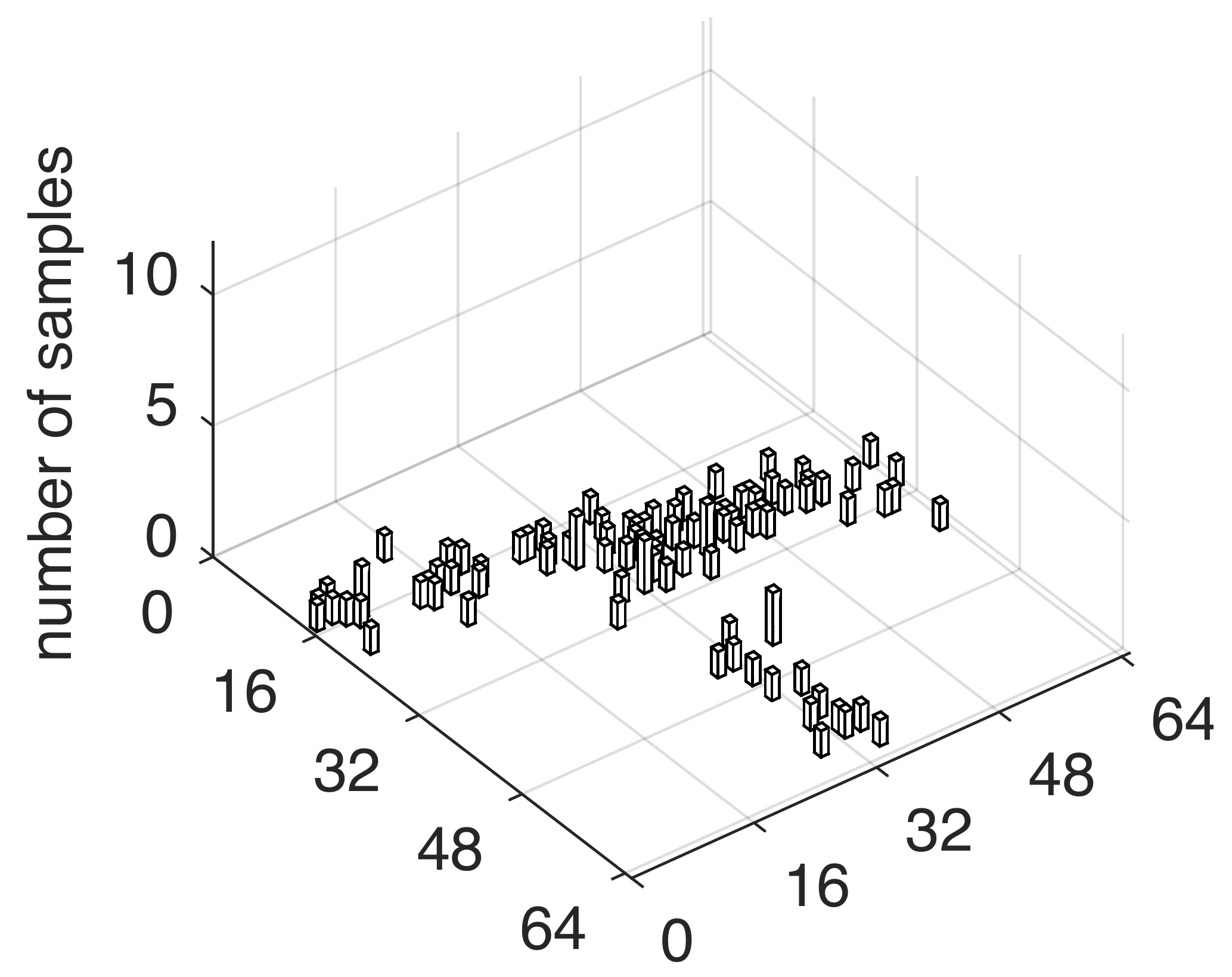}
}
\\ \vspace{-2ex}
\subfloat[Sub-resolution\label{fig:aug:res}]{
\includegraphics[width=0.49\columnwidth]{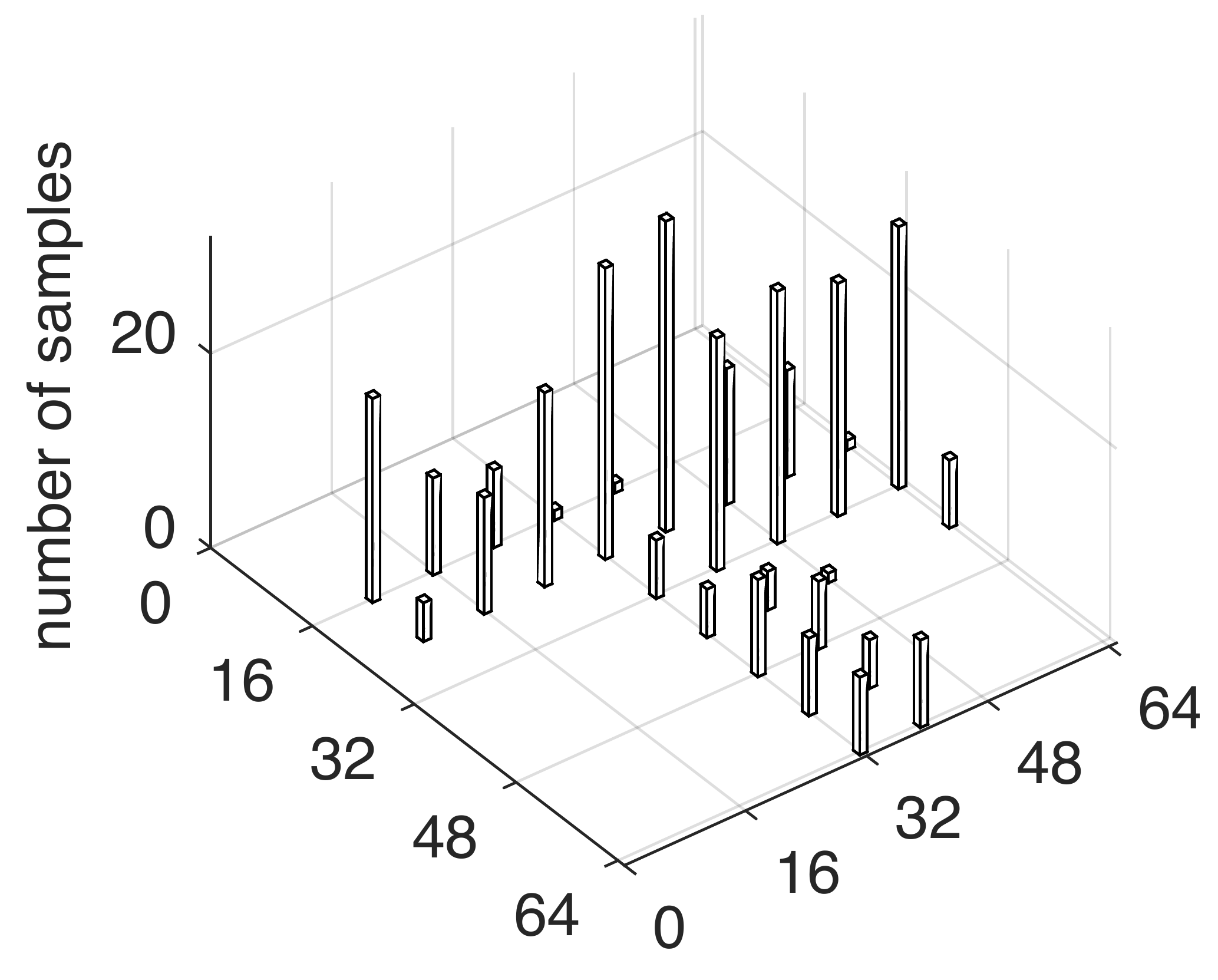}
}
\subfloat[Random omission\label{fig:aug:omission}]{
\includegraphics[width=0.49\columnwidth]{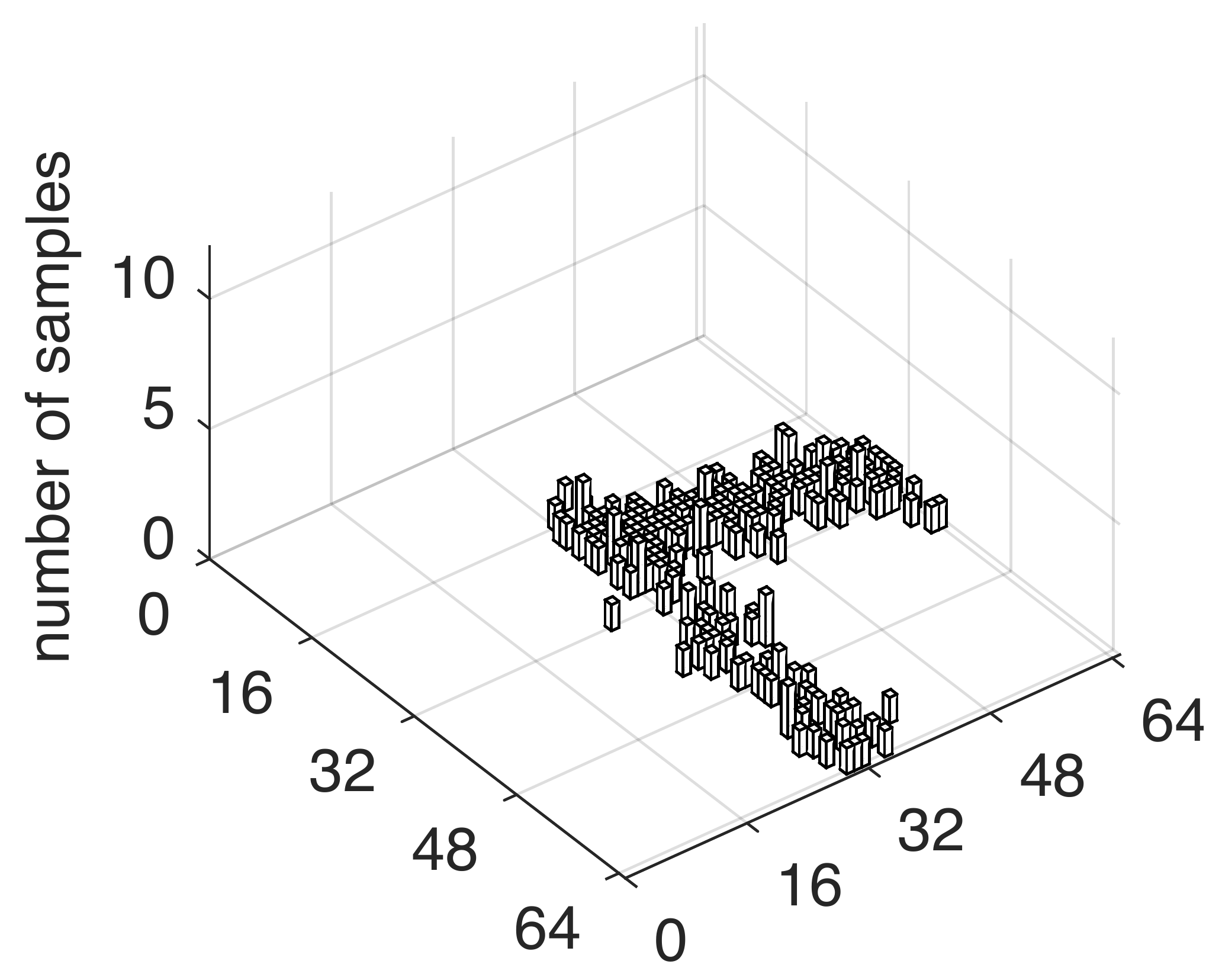}
}
\caption{GPS data augmentation}\label{fig:augment-methods}
\end{figure}

Deep CNNs are very complex models prone to overfitting, especially for GPS data that is relatively simple and well structured.
In our experiments, adding a GPS layer without any data augmentation leads to a superficial model which enhances RGB-based predictions wherever GPS data is dense, and suppresses the prediction wherever there is no GPS data.
In addition, the model is very sensitive to GPS quantity and quality.
For example, if we remove the GPS input altogether, the prediction is a lot worse than the model trained with RGB image input only.
We develop the following augmentation methods to prevent overfitting.
\begin{itemize}
	\setlength\itemsep{1pt}
	\item Randomly subsample the input GPS data
	\item Reduce the resolution of the input GPS data by a random factor, called sub-resolution hereafter
	\item Random perturbation of the GPS data
	\item Omitting a random area of GPS data
\end{itemize}
Figure~\ref{fig:augment-methods} illustrates some of these augmentation techniques.
Figure~\ref{fig:aug:original} shows the GPS samples on a 64 x 64 image patch.
The height of the bars indicates the number of samples projected to the same pixel, between zero to three in this case.
Figure~\ref{fig:aug:subsample} takes a random 60\% of samples from Fig.~\ref{fig:aug:original}.
Figure~\ref{fig:aug:res} reduces all samples to 1/8 of their original resolution such that the samples are aggregated to a small set of pixels.
Many GPS data have low resolution either because of inferior GPS receivers used or because of privacy protection.
In addition, sub-resolution leads to much higher values for the remaining pixels than the original data, which is similar to the case of larger GPS quantities.
The model trained with sub-resolution handles unseen larger amount of GPS data better in our experiments.
Figure~\ref{fig:aug:omission} omits samples on the left 32 x 32 square.

\subsection{Rendering} \label{sec:rendering}
\begin{figure}[!ht]
\center
  \vspace{-4ex}
\subfloat[Linear scale\label{fig:kernel:linear}]{
\includegraphics[width=0.4\columnwidth]{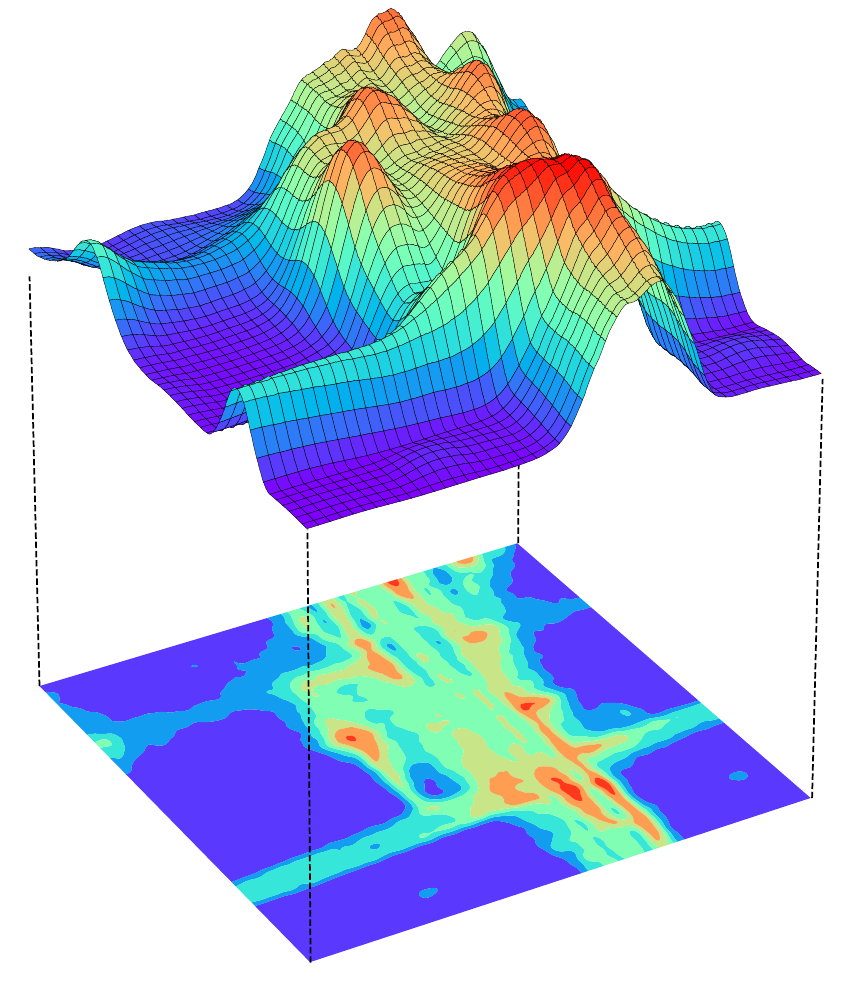}}
\subfloat[Log scale\label{fig:kernel:log}]{
\includegraphics[width=0.4\columnwidth]{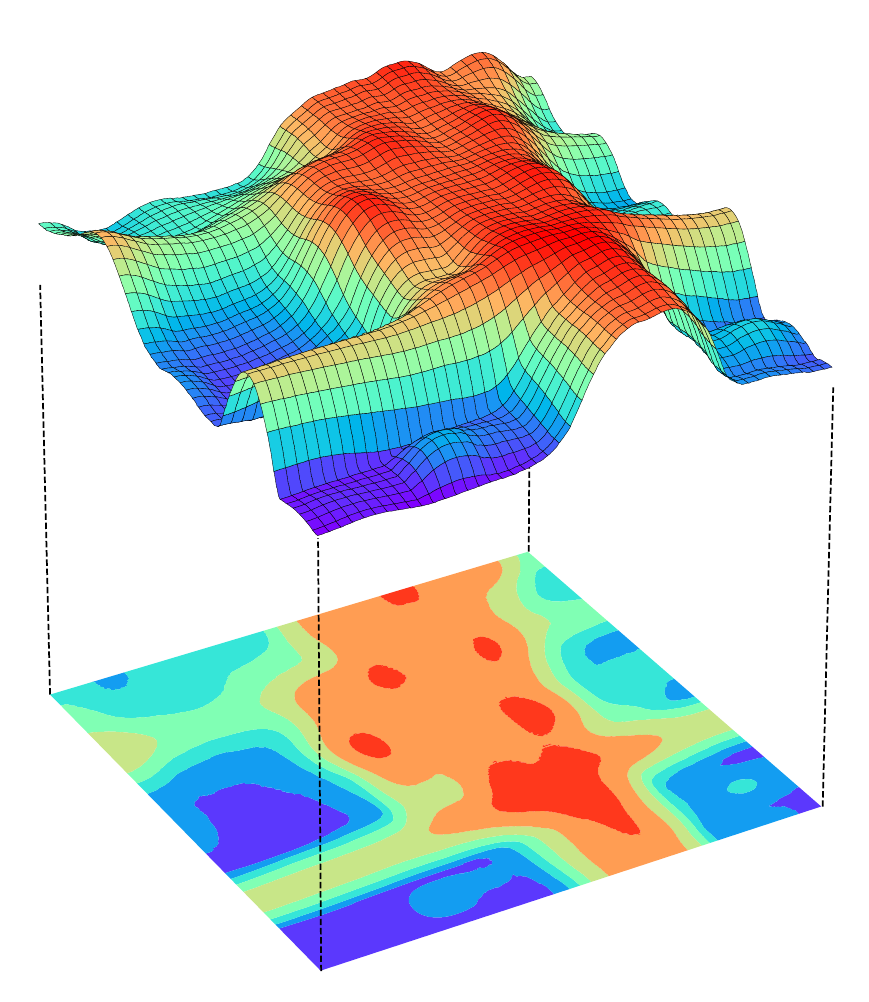}
}
\caption{Gaussian kernel rendering of Fig.~\ref{fig:gps:a}}
\label{fig:kernel-render}
  \vspace{-1ex}
\end{figure}

After augmentation, we must render the GPS data as an image layer to concatenate with the RGB image input.
There are many different ways to render the image.
For example in Fig.~\ref{fig:gps-noise}, we render a pixel white if and only if there is at least one GPS sample projected to it.
This method works with small datasets only.
As the GPS quantity increases, noise spreads and too many pixels will be white, like Fig.~\ref{fig:gps:a}.

Instead of a binary image, we can use a greyscale image where the number at each pixel indicates the number of samples projected to it, therefore road pixels will have higher values than noise pixels as the quantity increases.
Inspired by Kernel Density Estimation~(KDE) frequently used in road inferencing from GPS data\cite{pervasive/DaviesBH06}, we can also render the GPS data with Gaussian kernel smoothing.
Figure~\ref{fig:kernel:linear} is the Gaussian kernel rendering of Fig.~\ref{fig:gps:a}.
Because of data disparity between highways and residential roads, log scale could make infrequently traveled roads more prominent.
For example in Fig.~\ref{fig:kernel:log}, the horizontal road at the bottom becomes much more visible than in the linear scale.

When there is a limited quantity of GPS data but the sampling frequency is high, adding a line segment between consecutive samples helps~\cite{liu2012mining}, which is another way to render GPS data.
In our case, these line segments often shortcut intersections and curves because of low sampling frequency, and therefore do not improve results in our experiments.
Our 1D decoder has similar effect at roads with sparse samples, and they are not affected by sampling intervals.

Other GPS measurements can be useful for road extraction.
We render these measurements as separate input layers.
More specifically, the pixel values of the interval, speed, and bearing layers are the average sampling interval, average speed, and average sinusoid of the bearing for all the samples projected to the pixel, respectively.

\section{Experiments}

We experiment with the satellite imagery and our GPS datasets from two cities, and report our results here.

\textbf{Datasets}~~~
For satellite imagery, we crawled 350 images in Beijing and 50 images in Shanghai from Gaode map~\cite{gaode}.
All these images are 1024 x 1024 in size and 50~cm/pixel in resolution, a total area of about 100~km$^2$.
Like the DeepGlobe dataset, we manually created the training labels by masking out road pixels in the images.
We choose the same input image size as the DeepGlobe data set for the convenience of comparison.
It is also an appropriate size because a smaller one would lose the context and a larger size may not fit in GPU memory.
The DeepGlobe dataset is for much larger areas but we do not have GPS data in the areas for experiments.
Some other research work used large datasets by rendering OSM road vectors with fixed width, typically for developed countries~\cite{MnihThesis,bastani2018roadtracer}.
Roads in developing countries vary in width more significantly, and misalignments are prevalent on OSM.
Therefore we have to label road pixels manually.
Nevertheless, our dataset is among the largest in research work that do not use DeepGlobe datasets or OSM labels~\cite{mattyus2017deeproadmapper,zhu2017deep}.

Our GPS datasets are taxi and bus samples that include timestamp, latitude, longitude, speed, bearing, and vehicle status flags.
As discussed in Section~\ref{sec:gps}, our GPS datasets are from different devices with varying sampling rates and different resolutions for the measurements.

Similar to the competition and the other research work, we use the intersection over union~(IoU) as the main evaluation criteria, and report the average IoU among all test image patches.
We randomly split our dataset into three partitions, 70\% for training,  10\% for validation, and the rest 20\% for testing.
Other than the last experiment that evaluates the ability for our model to predict new areas, we use only the Beijing satellite images and GPS dataset for training and testing.

\begin{table}[!tbp]
\center
\small
\begin{threeparttable}
  \caption{Different input and model combinations}
  \label{tab:models}
  \vspace{-1ex}
  \begin{tabular}{|c|c|cc|}
    \hline
    
    \multirow{2}{*}{input}&\multirow{2}{*}{method}&
    \multicolumn{2}{c|}{IoU (\%) on \textit{test} set}\cr
    \cline{3-4}
    & & \quad plain \quad & 1D decoder\\
    \hline
    \hline
    \multirow{6}*{GPS}   & KDE~\cite{pervasive/DaviesBH06} & 34.06 & -\\
    & DeepLab (v3+)~\cite{chen2018deeplab}& 47.65  & - \\
    & U-Net~\cite{ronneberger2015u} &  43.63 & 48.10  \\
    & Res U-Net~\cite{zhang2017road} & 45.33 & 48.52 \\
    & LinkNet~\cite{chaurasia2017linknet} & 49.98 & \textbf{51.06} \\
    & D-LinkNet~\cite{zhou2018d}  & 48.46 & 49.95 \\
    \hline
    	\multirow{5}*{image} & DeepLab (v3+) & 43.40 & - \\
       	& U-Net & 51.85  & 52.10  \\
    	& Res U-Net & 50.26 & 51.77  \\
 		& LinkNet & 53.96 & 54.84 \\
 		& D-LinkNet & 54.42  & \textbf{55.15} \\
	\hline
	\multirow{5}*{image + GPS} & DeepLab (v3+) &  50.81  & -\\
		& U-Net &  53.22 & 54.88  \\
		& Res U-Net & 52.29  & 54.24 \\
		& LinkNet & 57.48  & 57.89 \\
		& D-LinkNet & 56.96  & \textbf{57.96} \\
    \hline
  \end{tabular}
\end{threeparttable}
\end{table}

\textbf{Models}~~~Our GPS rendering method applies to all existing segmentation models.
Here we choose DeepLab, two variants of U-Net, and two variants of LinkNet to evaluate.
The two variants of U-Net are the original one and the one with ResNet style encoder and decoder, denoted as Res U-Net. 
The two variants of LinkNet are the original one and D-LinkNet that achieved top performance in the DeepGlobe challenge.
For road extraction using GPS input only, we also add KDE method for comparison since it was among the best using traditional machine learning techniques~\cite{liu2012mining}.

\textbf{Baseline}~~~Our first experiment takes the GPS input alone; see the top section of Table~\ref{tab:models}.
For the KDE method, since we measure IoU only and do not extract road centerlines, we simply pick the best kernel size and the threshold to binarize the Gaussian smoothed image.
The results show that deep neural nets perform much better than the KDE method to extract roads from GPS data only.
Our 1D decoder is very useful against relatively shallow neural nets, and give about 1~\% increase against more complex models.
LinkNet shows the best result here.
Although D-LinkNet performed better in the DeepGlobe challenge, its additional complexity over LinkNet leads to more severe overfitting of the relatively simple GPS data.
We do not apply 1D decoder to DeepLab since it uses a bi-linear interpolation decoder without any transpose convolution.

Next we examine the performance of the different segmentation models with the satellite image input only; see the second section of Table~\ref{tab:models}.
The result is consistent with the numbers reported in the DeepGlobe challenge, where D-LinkNet is slightly better than the other models~\cite{zhou2018d}.
The best IoU in our test is lower than the number in the challenge because the Beijing area is more challenging than the rural areas and towns used in the challenge.
Many roads are completely blocked by tree canopy in the old city center area, and the road boundaries are not easy to define with the prevalent express/local/bike way systems.
DeepLab has the worst performance among the models we use.
Visually examining the output reveals much coarser borders than in the other model output, likely due to the bi-linear interpolation decoder instead of transpose convolution used.

Finally, with both the image and the GPS input, D-LinkNet remains the top performer.
Here the largest performance gain for the additional GPS input is DeepLab.
For the other models that already perform relatively well on the image input, the performance gain is about 2~\%, and the 1D decoder adds about another 1~\%.

\begin{figure}[!htbp]
\center
\subfloat[Performance with different GPS quantity\label{fig:augment:a}]{
\includegraphics[width=1\linewidth]{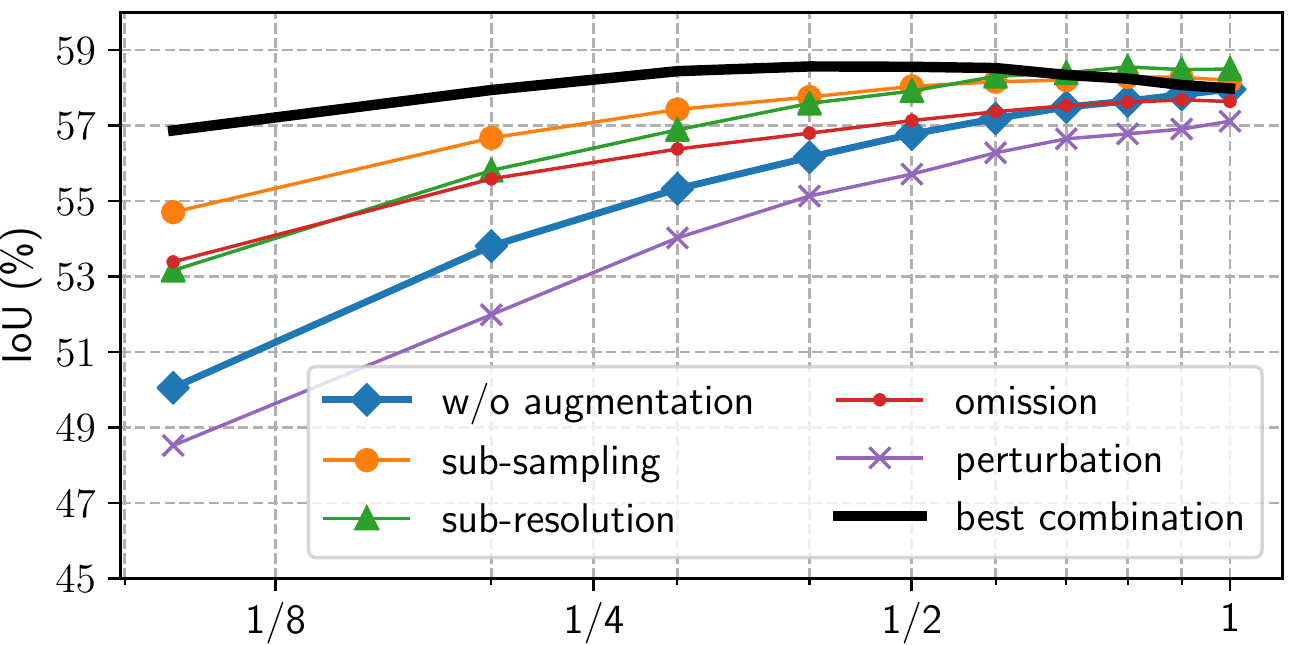}
}\\
\vspace{-3pt}
\subfloat[Performance with different GPS resolution \label{fig:augment:b}]{
\includegraphics[width=1\linewidth]{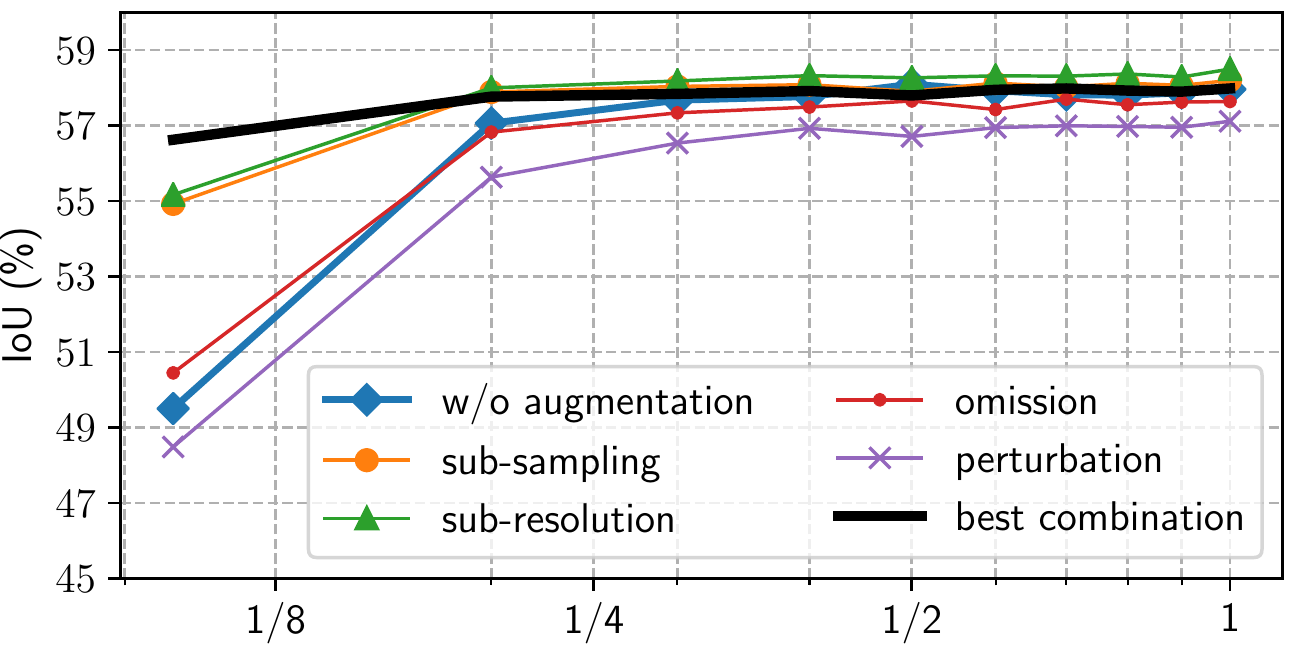}
}
\vspace{-1pt}
\caption{GPS data augmentation results (D-LinkNet with image+GPS input)}
\label{fig:augment}
\vspace{-3pt}
\end{figure}

\textbf{Augmentation}~~~Figure~\ref{fig:augment} shows the effectiveness of our GPS data augmentation.
Figure~\ref{fig:augment:a} and Figure~\ref{fig:augment:b} are the performance of different augmentation techniques with a subset of input data and a reduced resolution of input data, respectively.
With our data augmentation, our model not only performs much better with degraded GPS data input, but also gains about 0.5\% over the top-of-the-line performance.

\begin{figure}[htbp]
\center
\includegraphics[width=0.9\linewidth]{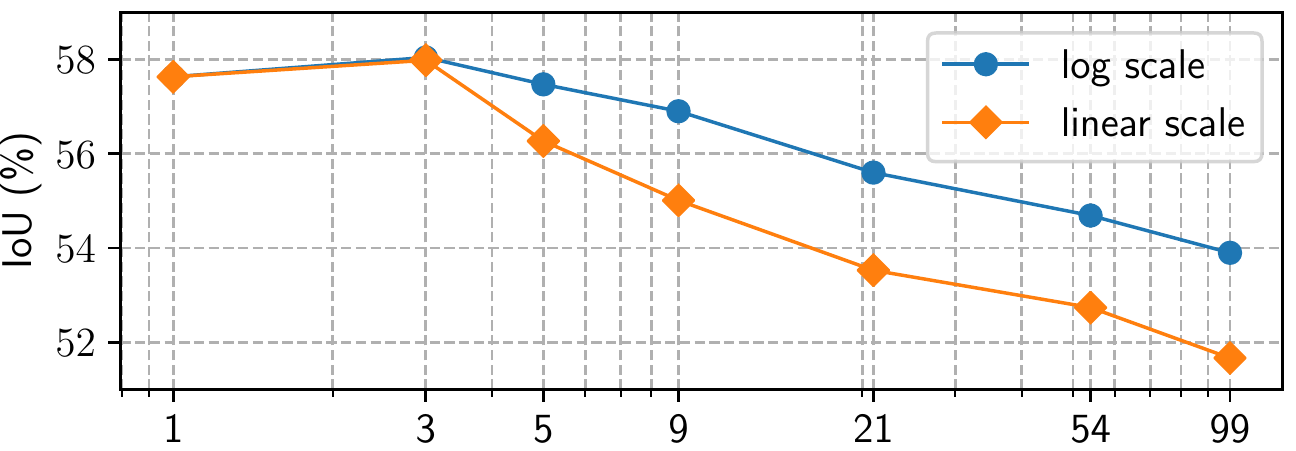}
\vspace{-1ex}
\caption{Rendering with different Gaussian kernel sizes}
\label{fig:kdesize}
\end{figure}

\textbf{Rendering}~~~As described in Section~\ref{sec:rendering}, Fig.~\ref{fig:kdesize} shows the performance of Gaussian kernel rendering with different kernel sizes and different rendering scale.
We also experimented with various combination of GPS measurements, sampling interval, vehicle speed, and vehicle bearing.
Adding another input layer of sampling interval alone gives the best performance gain.
Based on these results, we use two input layers for the GPS data for the rest of the experiments, Gaussian kernel rendering of the GPS samples with kernel size three and the sampling interval channel.

Table~\ref{tab:overall} is the overall performance gain with various improvements over the baseline using the image input only.
Altogether we achieved 4.76\% performance gain.

\begin{table}[htbp]
\small
\center
  \caption{Using GPS features and data augmentation}
  \label{tab:overall}
  \vspace{-1ex}
  \begin{tabular}{|l|c|}
    \hline
 settings (all using D-LinkNet) & IoU (\%) \\
    \hline
    \hline
    image                    & 54.42 \\
  image + GPS                       & 56.96 \\
  \hline
  image + GPS + 1D decoder                & 57.96 \\
  image + GPS + 1D decoder + augment.        & 58.55 \\
  \hline
 image + GPS + interval + 1D decoder          &  58.55 \\
 image + GPS + interval + 1D decoder + augment.  & \textbf{59.18} \\
    \hline
  \end{tabular}
  \vspace{-1ex}
\end{table}

\begin{figure}[!hbpt]
\center
\subfloat[GPS samples over satellite image\label{fig:improve:sat}]{
\includegraphics[width=0.49\linewidth]{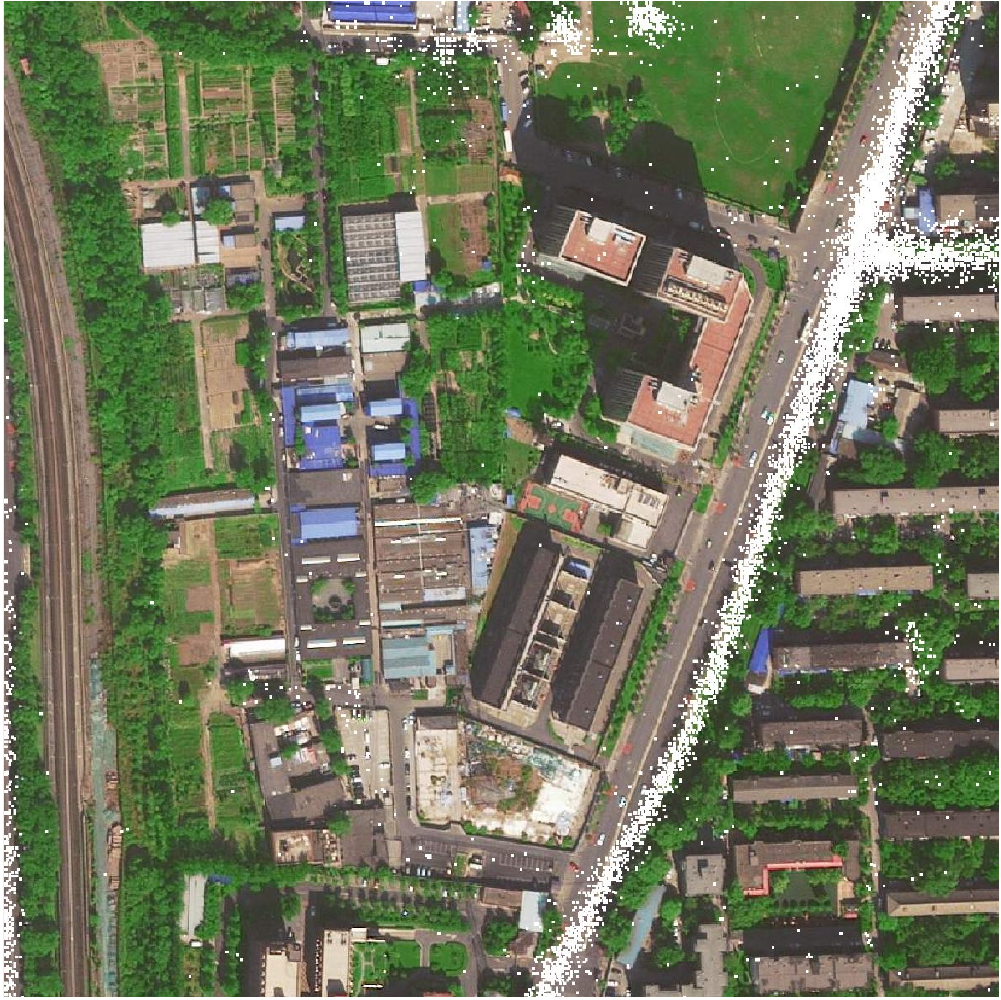}}
\subfloat[Roads confirmed by GPS\label{fig:improve:pred}]{
\includegraphics[width=0.49\linewidth]{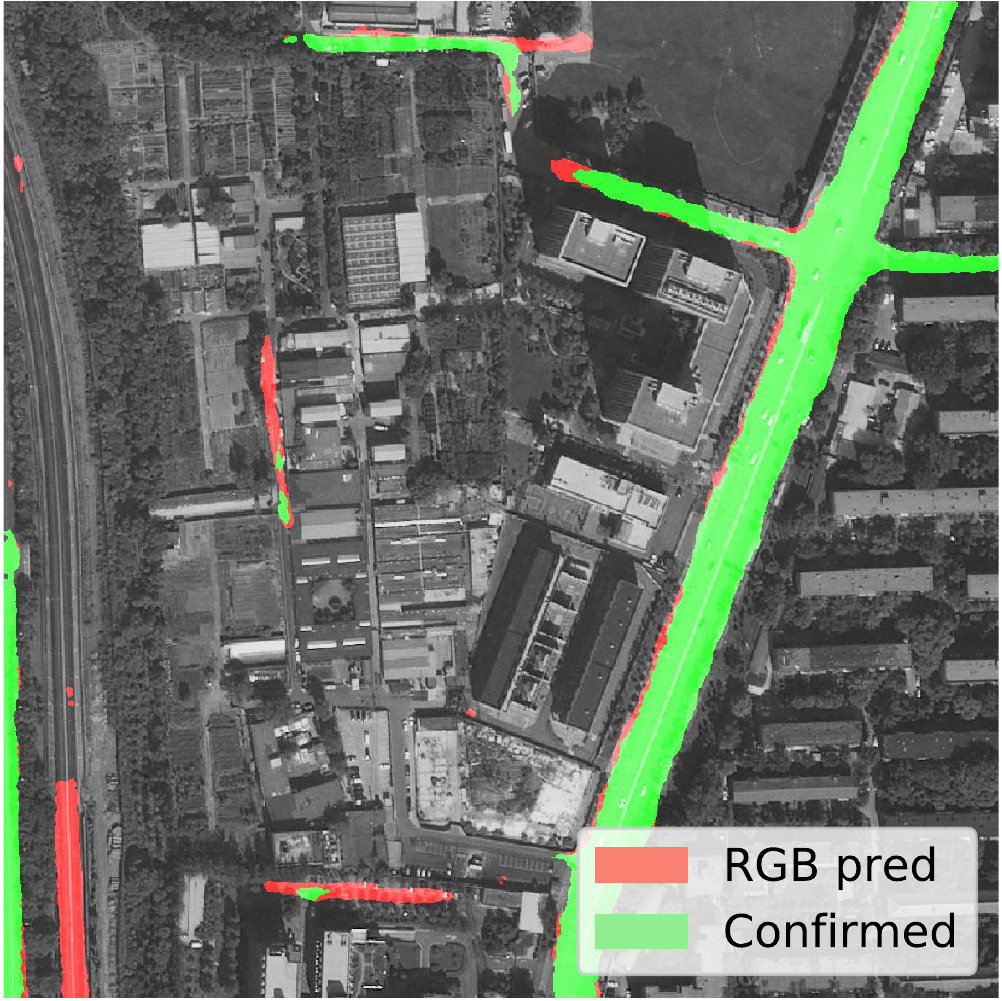}}
\vspace{-1ex}
\caption{Road verification using GPS data}
\label{fig:confirm}
\end{figure}

\textbf{GPS as verification}~~~As discussed in Section~\ref{sec:intro}, local verification is often required for mapping.
Figure~\ref{fig:confirm} shows how the crowdsourced GPS data can serve the verification purpose without local survey.
Here the green pixels are high confidence predictions by both the image-only input and the image + GPS input, while red pixels are high confidence predictions by image-only input but low confidence predictions when GPS input is added.
Map matching could give additional confidence by matching GPS traces to roads by topology~\cite{gis/NewsonK09}, which is beyond the scope of this paper. 

\begin{figure*}
\center
\subfloat{
\includegraphics[width=0.99\linewidth]{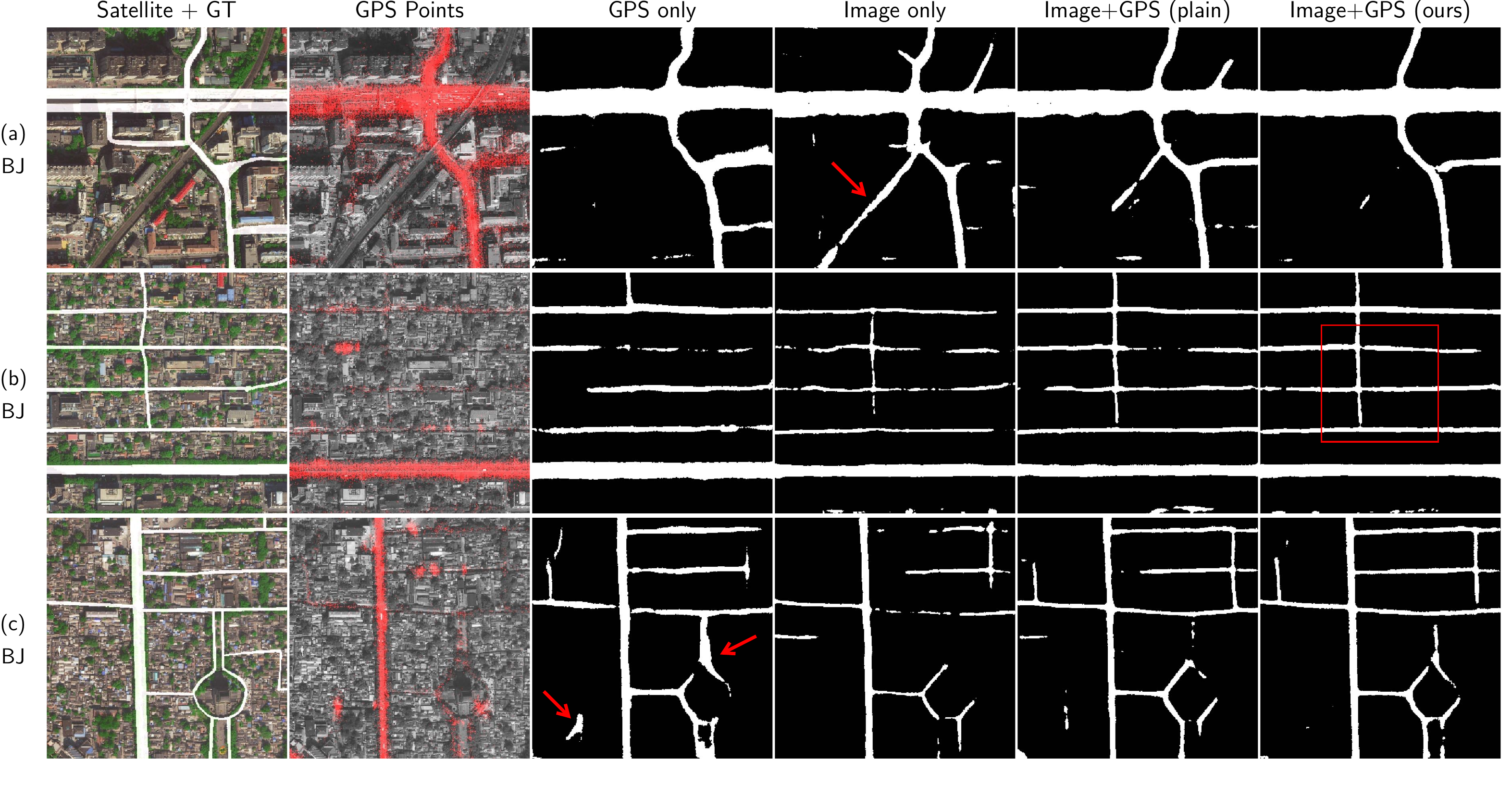}}\\
\vspace{-1.5em}
\subfloat{
\includegraphics[width=0.99\linewidth]{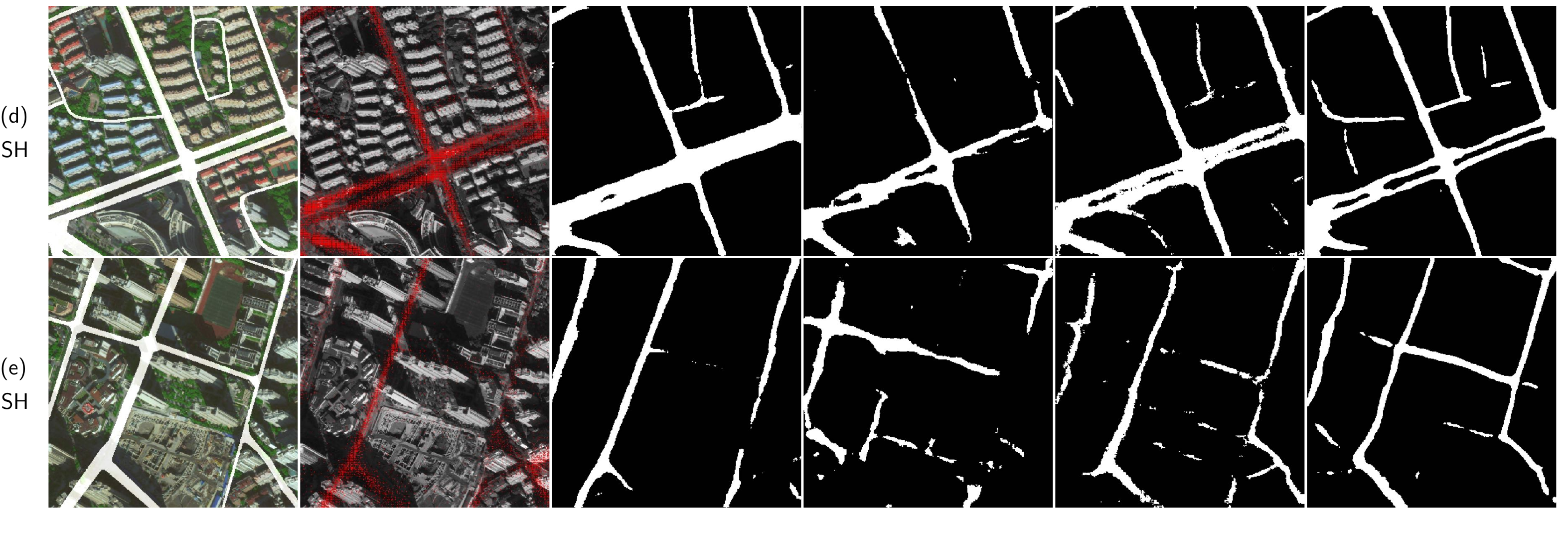}}
\vspace{-3ex}
\caption{Prediction results using different methods on Beijing and Shanghai testing datasets trained on Beijing dataset only}
\vspace{-2ex}
\label{fig:result}
\end{figure*}

\textbf{New testing area}~~~Table~\ref{tab:prednew} is the testing results with our Shanghai dataset using different training data and methods.
Despite the different GPS data characteristics, it is evident that prediction with additional GPS input is more resilient in the new domain, 18.9\% IoU drop for the model trained with both datasets instead of 31.6\% for the model trained with image input only.
The performance gain is enhanced when employing the GPS data augmentation, confirming its effect against overfitting.
\begin{table}[!htbp]
\small
\center
  \vspace{-1ex}
  \caption{Shanghai testing dataset results}
  \label{tab:prednew}
  \vspace{-1ex}
  \begin{tabular}{|c|c|c|c|}
    \hline
    train & method & IoU(\%) & relative \\
    \hline
    \hline
     \multirow{4}*{\specialcell{Beijing\\+\\Shanghai}}  & GPS &  44.88 & --\\
    &  image & 55.76 & -- \\
   & \specialcell{image + GPS (w/o augment)} & 59.30 & --  \\
   & \specialcell{image + GPS (w/ augment)}& \textbf{60.00} & -- \\
    \hline
	\multirow{4}*{Beijing} & GPS & 42.82  & -4.6\%\\
    & image & 38.16 & -31.6\% \\
   	&	 \specialcell{image + GPS (w/o augment)} & 44.57 & -24.9\%  \\
   	&   \specialcell{image + GPS (w/ augment)}& \textbf{48.69} &-18.9\%  \\
    \hline
  \end{tabular}
\end{table}

\textbf{Qualitative results}~~~Figure \ref{fig:result} visualizes the road extraction results of different methods in different testing areas of Beijing and Shanghai, trained using Beijing dataset only.
Overall, prediction using GPS data only largely matches the sample distribution.
With the image input only, occlusion and other image issues can cause poor performance.
Both image and GPS input give the best results, and our enhancement techniques give a bit cleaner output.
As examples, the areas pointed by red arrows show false positives removed with our model using GPS data.
The one in the first row is a railway and the one in the third row is from GPS noise.
The red square shows an area with dense tree canopy and relatively sparse GPS samples.
Only the combination of image and GPS data extracts a relatively complete road network.

\section{Conclusion}
With large-scale crowdsourced GPS datasets, fusing GPS data with aerial image input gives much better road segmentation results than using images or GPS data alone with deep neural net models.
Inspired by image augmentation techniques, our GPS data augmentation is very effective against overfitting, and thus our method performs much better in new testings areas than other models.
In our experiences, aerial imagery works best for residential roads detection because they are relatively simple, numerous, and infrequently traveled.
In contrast, GPS data can recover arterial roads with ease even for complicated highway systems and under severe image occlusion.
Therefore, the two data sources well complement each other for road extraction tasks.

\section*{Acknowledgement}
We thank anonymous reviewers for valuable feedback.  This research is supported by NSFC General Program 41771481, Shanghai Science and Technology Commission program 17511104502, and a gift fund from Facebook.

\vspace{1cm}

{\small
\bibliographystyle{ieee}
\bibliography{cvpr19}

\begin{thebibliography}{10}\itemsep=-1pt

\bibitem{OSM_AI_assisted}
{AI}-assisted road tracing.
\newblock \url{wiki.openstreetmap.org/wiki/AI-Assisted_Road_Tracing}.

\bibitem{gaode}
{Gaode Map}.
\newblock \url{www.amap.com}.

\bibitem{OSM}
{OpenStreetMap}.
\newblock \url{www.openstreetmap.org}.

\bibitem{aksoy2008performance}
S.~Aksoy, B.~Ozdemir, S.~Eckert, F.~Kayitakire, M.~Pesarasi, O.~Aytekin, C.~C.
  Borel, J.~Cech, E.~Christophe, S.~Duzgun, et~al.
\newblock Performance evaluation of building detection and digital surface
  model extraction algorithms: Outcomes of the prrs 2008 algorithm performance
  contest.
\newblock In {\em IAPR Workshop on Pattern Recognition in Remote Sensing}.
  IEEE, 2008.

\bibitem{bastani2018roadtracer}
F.~Bastani, S.~He, S.~Abbar, M.~Alizadeh, H.~Balakrishnan, S.~Chawla,
  S.~Madden, and D.~DeWitt.
\newblock Roadtracer: Automatic extraction of road networks from aerial images.
\newblock In {\em Computer Vision and Pattern Recognition (CVPR)}, 2018.

\bibitem{gis/BiagioniE12}
J.~Biagioni and J.~Eriksson.
\newblock Map inference in the face of noise and disparity.
\newblock In {\em SIGSPATIAL Conference on Geographic Information Systems
  (GIS)}, 2012.

\bibitem{chaurasia2017linknet}
A.~Chaurasia and E.~Culurciello.
\newblock Linknet: Exploiting encoder representations for efficient semantic
  segmentation.
\newblock In {\em Visual Communications and Image Processing (VCIP)}, 2017.

\bibitem{chen2018deeplab}
L.-C. Chen, G.~Papandreou, I.~Kokkinos, K.~Murphy, and A.~L. Yuille.
\newblock Deeplab: Semantic image segmentation with deep convolutional nets,
  atrous convolution, and fully connected crfs.
\newblock {\em IEEE transactions on pattern analysis and machine intelligence},
  40(4):834--848, 2018.

\bibitem{pervasive/DaviesBH06}
J.~J. Davies, A.~R. Beresford, and A.~Hopper.
\newblock Scalable, distributed, real-time map generation.
\newblock {\em IEEE Pervasive Computing}, 5(4):47--54, 2006.

\bibitem{DeepGlobe18}
I.~Demir, K.~Koperski, D.~Lindenbaum, G.~Pang, J.~Huang, S.~Basu, F.~Hughes,
  D.~Tuia, and R.~Raskar.
\newblock Deepglobe 2018: A challenge to parse the earth through satellite
  images.
\newblock In {\em Computer Vision and Pattern Recognition (CVPR) Workshops},
  2018.

\bibitem{huang2018large}
B.~Huang, K.~Lu, N.~Audebert, A.~Khalel, Y.~Tarabalka, J.~Malof, A.~Boulch,
  B.~Le~Saux, L.~Collins, K.~Bradbury, et~al.
\newblock Large-scale semantic classification: outcome of the first year of
  inria aerial image labeling benchmark.
\newblock In {\em IEEE International Geoscience and Remote Sensing Symposium
  (IGARSS)}, 2018.

\bibitem{karagiorgou2017layered}
S.~Karagiorgou, D.~Pfoser, and D.~Skoutas.
\newblock A layered approach for more robust generation of road network maps
  from vehicle tracking data.
\newblock {\em ACM Transactions on Spatial Algorithms and Systems (TSAS)},
  3(1):3, 2017.

\bibitem{lecun1998gradient}
Y.~LeCun, L.~Bottou, Y.~Bengio, and P.~Haffner.
\newblock Gradient-based learning applied to document recognition.
\newblock {\em Proceedings of the IEEE}, 86(11):2278--2324, 1998.

\bibitem{li2016road}
P.~Li, Y.~Zang, C.~Wang, J.~Li, M.~Cheng, L.~Luo, and Y.~Yu.
\newblock Road network extraction via deep learning and line integral
  convolution.
\newblock In {\em IEEE International Geoscience and Remote Sensing Symposium
  (IGARSS)}, 2016.

\bibitem{liu2012mining}
X.~Liu, J.~Biagioni, J.~Eriksson, Y.~Wang, G.~Forman, and Y.~Zhu.
\newblock Mining large-scale, sparse gps traces for map inference: comparison
  of approaches.
\newblock In {\em ACM SIGKDD international conference on Knowledge discovery
  and data mining}, 2012.

\bibitem{mattyus2017deeproadmapper}
G.~M{\'a}ttyus, W.~Luo, and R.~Urtasun.
\newblock Deeproadmapper: Extracting road topology from aerial images.
\newblock In {\em The IEEE International Conference on Computer Vision (ICCV)},
  2017.

\bibitem{mayer2006test}
H.~Mayer, S.~Hinz, U.~Bacher, and E.~Baltsavias.
\newblock A test of automatic road extraction approaches.
\newblock {\em International Archives of Photogrammetry, Remote Sensing, and
  Spatial Information Sciences}, 36(3):209--214, 2006.

\bibitem{MnihThesis}
V.~Mnih.
\newblock {\em Machine Learning for Aerial Image Labeling}.
\newblock PhD thesis, University of Toronto, 2013.

\bibitem{mnih2012learning}
V.~Mnih and G.~E. Hinton.
\newblock Learning to label aerial images from noisy data.
\newblock In {\em International conference on machine learning (ICML)}, 2012.

\bibitem{mosinska2018beyond}
A.~J. Mosinska, P.~Marquez~Neila, M.~Kozinski, and P.~Fua.
\newblock Beyond the pixel-wise loss for topology-aware delineation.
\newblock In {\em Computer Vision and Pattern Recognition (CVPR)}, 2018.

\bibitem{gis/NewsonK09}
P.~Newson and J.~Krumm.
\newblock Hidden markov map matching through noise and sparseness.
\newblock In {\em SIGSPATIAL Conference on Geographic Information Systems
  (GIS)}, 2009.

\bibitem{STOM:Facebook:2018}
D.~Patel.
\newblock Osm at facebook.
\newblock State of the Map, 2018.

\bibitem{kdd/RogersLW99}
S.~Rogers, P.~Langley, and C.~Wilson.
\newblock Mining gps data to augment road models.
\newblock In {\em ACM SIGKDD international conference on Knowledge discovery
  and data mining}, 1999.

\bibitem{ronneberger2015u}
O.~Ronneberger, P.~Fischer, and T.~Brox.
\newblock U-net: Convolutional networks for biomedical image segmentation.
\newblock In {\em International Conference on Medical image computing and
  computer-assisted intervention}, pages 234--241. Springer, 2015.

\bibitem{SchrodlWRLW04}
S.~Schr{\"o}dl, S.~Schr{\"o}dl, K.~Wagstaff, S.~Rogers, P.~Langley, and
  C.~Wilson.
\newblock Mining {GPS} traces for map refinement.
\newblock {\em Data Mining Knowledge Discovery}, 9(1):59--87, 2004.

\bibitem{shan2015cobweb}
Z.~Shan, H.~Wu, W.~Sun, and B.~Zheng.
\newblock Cobweb: a robust map update system using gps trajectories.
\newblock In {\em ACM International Joint Conference on Pervasive and
  Ubiquitous Computing (UbiComp)}, 2015.

\bibitem{shelhamer2017fully}
E.~Shelhamer, J.~Long, and T.~Darrell.
\newblock Fully convolutional networks for semantic segmentation.
\newblock {\em IEEE transactions on pattern analysis and machine intelligence},
  39(4):640--651, 2017.

\bibitem{sun2018combining}
T.~Sun, Z.~Di, and Y.~Wang.
\newblock Combining satellite imagery and gps data for road extraction.
\newblock 2018.
\newblock SIGSPATIAL Conference on Geographic Information Systems (GIS)
  workshop.

\bibitem{STOM:Microsoft:2018}
M.~Trifunovic.
\newblock Robot tracers - extraction and classification at scale using \& cntk.
\newblock State of the Map, 2018.

\bibitem{Diggelen07}
F.~van Diggelen.
\newblock Gnns accuracy: Lies, damn lies, and statistics.
\newblock {\em GPS World}, pages 26--32, 2007.

\bibitem{volpi2017dense}
M.~Volpi and D.~Tuia.
\newblock Dense semantic labeling of subdecimeter resolution images with
  convolutional neural networks.
\newblock {\em IEEE Transactions on Geoscience and Remote Sensing},
  55(2):881--893, 2017.

\bibitem{wang2017torontocity}
S.~Wang, M.~Bai, G.~Mattyus, H.~Chu, W.~Luo, B.~Yang, J.~Liang, J.~Cheverie,
  S.~Fidler, and R.~Urtasun.
\newblock Torontocity: Seeing the world with a million eyes.
\newblock In {\em International Conference on Computer Vision (ICCV)}, 2017.

\bibitem{Wang:2016keynote}
Y.~Wang.
\newblock {Scaling Maps at Facebook}.
\newblock In {\em SIGSPATIAL Conference on Geographic Information Systems
  (GIS)}, 2016.
\newblock keynote.

\bibitem{Wang:2013}
Y.~Wang, X.~Liu, H.~Wei, G.~Forman, C.~Chen, and Y.~Zhu.
\newblock Crowdatlas: Self-updating maps for cloud and personal use.
\newblock In {\em International Conference on Mobile Systems, Applications, and
  Services (MobiSys)}, 2013.

\bibitem{xia2018dota}
G.-S. Xia, X.~Bai, J.~Ding, Z.~Zhu, S.~Belongie, J.~Luo, M.~Datcu, M.~Pelillo,
  and L.~Zhang.
\newblock Dota: A large-scale dataset for object detection in aerial images.
\newblock In {\em Computer Vision and Pattern Recognition (CVPR)}, 2018.

\bibitem{yokoya2018open}
N.~Yokoya, P.~Ghamisi, J.~Xia, S.~Sukhanov, R.~Heremans, I.~Tankoyeu,
  B.~Bechtel, B.~Le~Saux, G.~Moser, and D.~Tuia.
\newblock Open data for global multimodal land use classification: Outcome of
  the 2017 ieee grss data fusion contest.
\newblock {\em IEEE Journal of Selected Topics in Applied Earth Observations
  and Remote Sensing}, 11(5):1363--1377, 2018.

\bibitem{yuan2016image}
J.~Yuan and A.~M. Cheriyadat.
\newblock Image feature based gps trace filtering for road network generation
  and road segmentation.
\newblock {\em Machine Vision and Applications}, 27(1):1--12, 2016.

\bibitem{zhang2017road}
Z.~Zhengxin, L.~Qingjie, and W.~Yunhong.
\newblock Road extraction by deep residual u-net.
\newblock In {\em IEEE GEOSCIENCE AND REMOTE SENSING LETTERS}, 2017.

\bibitem{zhou2018d}
L.~Zhou, C.~Zhang, and M.~Wu.
\newblock D-linknet: Linknet with pretrained encoder and dilated convolution
  for high resolution satellite imagery road extraction.
\newblock In {\em Computer Vision and Pattern Recognition (CVPR) Workshops},
  2018.

\bibitem{zhu2017deep}
X.~X. Zhu, D.~Tuia, L.~Mou, G.-S. Xia, L.~Zhang, F.~Xu, and F.~Fraundorfer.
\newblock Deep learning in remote sensing: a comprehensive review and list of
  resources.
\newblock {\em IEEE Geoscience and Remote Sensing Magazine}, 5(4):8--36, 2017.

\end{thebibliography}
}

\end{document}